\documentclass[10 pt, journal, published]{IEEEtran}

\usepackage{amsmath,amsfonts}
\usepackage{algorithm}
\usepackage{algorithmicx}
\usepackage{algpseudocode}
\usepackage{array}
\usepackage{textcomp}
\usepackage{stfloats}
\usepackage{subfigure}
\usepackage{url}
\usepackage{verbatim}
\usepackage{graphicx}
\usepackage{cite}
\usepackage{amssymb}
\usepackage{graphicx}
\usepackage{textcomp}
\usepackage{xcolor}
\usepackage{gensymb}
\usepackage[export]{adjustbox}
\usepackage{todonotes}
\usepackage{comment}
\usepackage{orcidlink}

\usepackage{hyperref}
\hypersetup{
    colorlinks=true,
    linkcolor=blue,
    filecolor=magenta,      
    urlcolor=cyan,
    pdftitle={Overleaf Example},
    pdfpagemode=FullScreen,
    }
    
\begin{document}

\title{Proximal Control of UAVs with Federated Learning for Human-Robot Collaborative Domains}

\author{Lucas Nogueira Nobrega, Ewerton de Oliveira \orcidlink{https://orcid.org/0000-0002-3160-3571}, Martin Saska \orcidlink{https://orcid.org/0000-0001-7106-3816},~\IEEEmembership{Member, IEEE}, and Tiago~Nascimento \orcidlink{https://orcid.org/0000-0002-9319-2114},~\IEEEmembership{Senior Member, IEEE}

\thanks{Received 20 May 2024; accepted 23 October 2024. Date of publication 4 November 2024; date of current version 11 November 2024. This paper was recommended for publication by Editor Gentiane Venture upon evaluation of the Associate Editor and Reviewers’ comments. \textit{(Corresponding author: Tiago Nascimento)}}
\thanks{This work has been supported by the National Council for Scientific and Technological Development – CNPq, by the National Fund for Scientific and Technological Development – FNDCT, and by the Ministry of Science, Technology and Innovations – MCTI from Brazil under research project No. 304551/2023-6 and 407334/2022-0, by the Paraiba State Research Support Foundation - FAPESQ under research project No. 3030/2021, by CTU grant no SGS23/177/OHK3/3T/13, by the Czech Science Foundation (GAČR) under research project No. 23-07517S, and by the Europen Union under the project Robotics and advanced industrial production (reg. no. $CZ.02.01.01/00/22\_008/0004590$).}

\thanks{Lucas Nogueira Nobrega and Ewerton de Oliveira are with the Lab of Systems Engineering and Robotics (LASER), Department of Computer Systems, Universidade Federal da Paraiba, João Pessoa 58051-900, Brazil.}
\thanks{Martin Saska is with the Department of Cybernetics, Czech Technical University in Prague, Prague 16629, Czechia (e-mail: {\tt\small martin.saska@fel.cvut.cz}).}
\thanks{Tiago Nascimento is with the Lab of Systems Engineering and Robotics (LASER), Department of Computer Systems, Universidade Federal da Paraiba, João Pessoa 58051-900, Brazil, and also with the Department of Cybernetics, Czech Technical University in Prague, Prague 16629, Czechia (e-mail: {\tt\small tiagopn@ci.ufpb.br}).}
\thanks{Digital Object Identifier (DOI): s10.1109/LRA.2024.3491417.}}

\markboth{IEEE ROBOTICS AND AUTOMATION LETTERS, PREPRINT VERSION. ACCEPTED OCT, 2024}%
{SANTOS \MakeLowercase{\textit{et al.}}: Proximal Control of UAVs with FL for HR Collaborative Domains}


\maketitle

\begin{abstract}
The human-robot interaction (HRI) is a growing area of research. In HRI, complex command (action) classification is still an open problem that usually prevents the real applicability of such a technique. The literature presents some works that use neural networks to detect these actions. However, occlusion is still a major issue in HRI, especially when using uncrewed aerial vehicles (UAVs), since, during the robot's movement, the human operator is often out of the robot's field of view. Furthermore, in multi-robot scenarios, distributed training is also an open problem. In this sense, this work proposes an action recognition and control approach based on Long Short-Term Memory (LSTM) Deep Neural Networks with two layers in association with three densely connected layers and Federated Learning (FL) embedded in multiple drones. The FL enabled our approach to be trained in a distributed fashion, i.e., access to data without the need for cloud or other repositories, which facilitates the multi-robot system's learning. Furthermore, our multi-robot approach results also prevented occlusion situations, with experiments with real robots achieving an accuracy greater than 96\%.
\end{abstract}

\begin{IEEEkeywords}
Aerial Systems: Application, UAV, Action Recognition, Human Detection and Tracking
\end{IEEEkeywords}

\section{Introduction}

\IEEEPARstart{R}{obots} have been proven useful in several dimensions of human life. They have, for example, fostered telepresence in critical environments by allowing humans to benefit from their mobility and bidirectional audio and video feeds~\cite{tsui_robots_2012}. The interaction between humans and robots often occurs through devices such as controllers and screens. Alternatively, other forms of control like wearable sensors (e.g., accelerometers), as well as in-palm and optical devices, have gained popularity~\cite{floersch_human_2021}. Fostering new ways of human-robot interface, embedded cameras have been used as a mainstream form of robot control via human gesture recognition. For Unmanned Aerial Vehicles (UAVs), also known as drones, this form of control is anchored on the fact that drones are increasingly present in our daily lives. For example, drones can be used to monitor and assist workers in their activities without requiring the use of physical controlling devices. This proximal form of control enables a seamless interaction experience where the human is free to perform other tasks requiring both
hands while the device becomes more likely to be perceived as a smart autonomous companion with an appropriate level of rationality~\cite{de2021deceptive}.

An example of a drone-assisted human task is described by Uzakov et al.~\cite{uzakov_uav_2020} where the authors have used three drones to follow a power line worker using computer vision to detect clothing of the human. In addition, Chaudhary et al. \cite{Akash2022} presented an action recognition approach to control a multi-robot system through vision. However, training multiple robots in a distributed fashion is still an open challenge. To overcome this problem, Federated Learning (FL) has been proposed in the literature \cite{XIANJIA2021135}.

Thus, in this work, we focus on experiments in the proximal control of UAVs in human-robot collaboration environments. We address the direct control of the drone via action recognition, taking into account the limitations of the platform w.r.t. computing, memory, and power management. For such a goal, we resort to FL~\cite{mcmahan2017communication}, a distributed learning paradigm that enables multiple devices to collaboratively train the Machine Learning-based action recognition model without sending the raw data out, improving experienced latency and relieving bandwidth and energy burden~\cite{qu_decentralized_2021}.

\section{Related Work}

As we mentioned above, we propose a Federated Learning and LSTM-based approach for proximal control of UAVs for human-robot interaction (HRI). To the best of our knowledge, this is the first work that uses Federated Learning in HRI for proximal control of UAVs. Therefore, in the literature, it is only possible to find papers with techniques that are only part of our contribution. For example, some works use Convolutional Neural Networks (CNN) and LSTM  only for static and dynamic gesture recognition. Works such as the one from Celebi et al. \cite{Celebi2013}, Rwigema et al. \cite{Rwigema2019}, and Hakim et al. \cite{hakim_dynamic_2019} focus on using CNN and LSTM to recognize more than 20 gestures (static and dynamic) for general purposes such as controlling a TV. Furthermore, specifically for action recognition of general purposes, Sozinov et al. \cite{sozinov_human_2018} proposed a federated learning approach to training a human activity recognition classifier to tackle the increase in communication costs and possible privacy infringement.

In addition, these techniques can also be used to control robots, which is the main application of our work. The literature presents some works that use CNN and LSTM to recognize such gestures and actions in order to control a single UAV \cite{cauchard_droneio_2019,kassab_real-time_2020,liu_real-time_2021}. However, when controlling robots, some problems are still open. For example, the type of the gesture (static or dynamic), the size and variance of the dataset, the robustness of the control approach, latency, and the multi-robot system cases are still open.

Thus, the first problem we aimed to tackle is the dataset for UAV control. The literature presents similar works, such as the one proposed by Kassab et al. \cite{kassab_real-time_2020} that created dataset and learning models for static gesture recognition to control one UAV. However, it is a small dataset and is limited to only hand and face recognition and only one UAV. In addition, Liu and Szirányi \cite{liu_real-time_2021} used a generic COCO dataset \cite{coco} that has more than 80 different objects to detect humans. Although this work now uses the full body for gesture and action recognition, the dataset is generic. In contrast, Perera et al. \cite{perera_uav-gesture_2019,perera_drone-action_2019} proposed a dataset of two static gestures and eleven actions. These gestures/actions are based on the general aircraft handling signals and helicopter handling signals to command a UAV. The dataset is available with body joints and gesture classes to be reusable, calculated by Openpose \cite{cao_openpose_2019}. However, in our tests, we also found that the size and variance of this last dataset are not sufficient for proximal control of UAVs.

A second problem we aimed to tackle is the human-swarm interaction, i.e., the use of human-robot interaction techniques to control a multi-robot system. The use of machine learning techniques to recognize features together with a multi-UAV system for general purposes is not new \cite{Liu_Yi_9184079, Chhikara_Prateek_9409140}. However, only recently that some preliminary works have used such approaches (e.g., k-Nearest Neighbor) to recognize actions and control a multi-UAV system \cite{Akash2022}. One of the problems that exist in a multi-UAV system controlled by HRI approaches is within the training process. Thus, with Federated Learning, we aim to be able to train our LSTM-based approach in a distributed fashion.

Thus, \textbf{this work proposes a solution for the human-swarm interaction problem, using computer vision in a machine-learning architecture with FL}. For that solution, we created experiments to compare the neural network performance between two different datasets and evaluate if the FL reduced the time spent by the swarm to detect an action. By using the proposed architecture, it is expected that personal data will be protected and also, with the sharing of learning between the clients, will be able to converge more rapidly \cite{mcmahan_communication-efficient_2017}. Furthermore, \textbf{this work also proposes improvements in the relationship between humans and drones, in which by implementing natural human interactions (NHI) as actions recognized through embedded machine learning, the robot can react to help the related human being}. Finally, we reaffirm that, to the best of our knowledge, this is the first work that uses FL in HRI for proximal control of UAVs. Thus, we aimed for this to be a preliminary but significant work.

\begin{figure}[tpb]
    \centerline{\includegraphics[width=1\linewidth, frame]{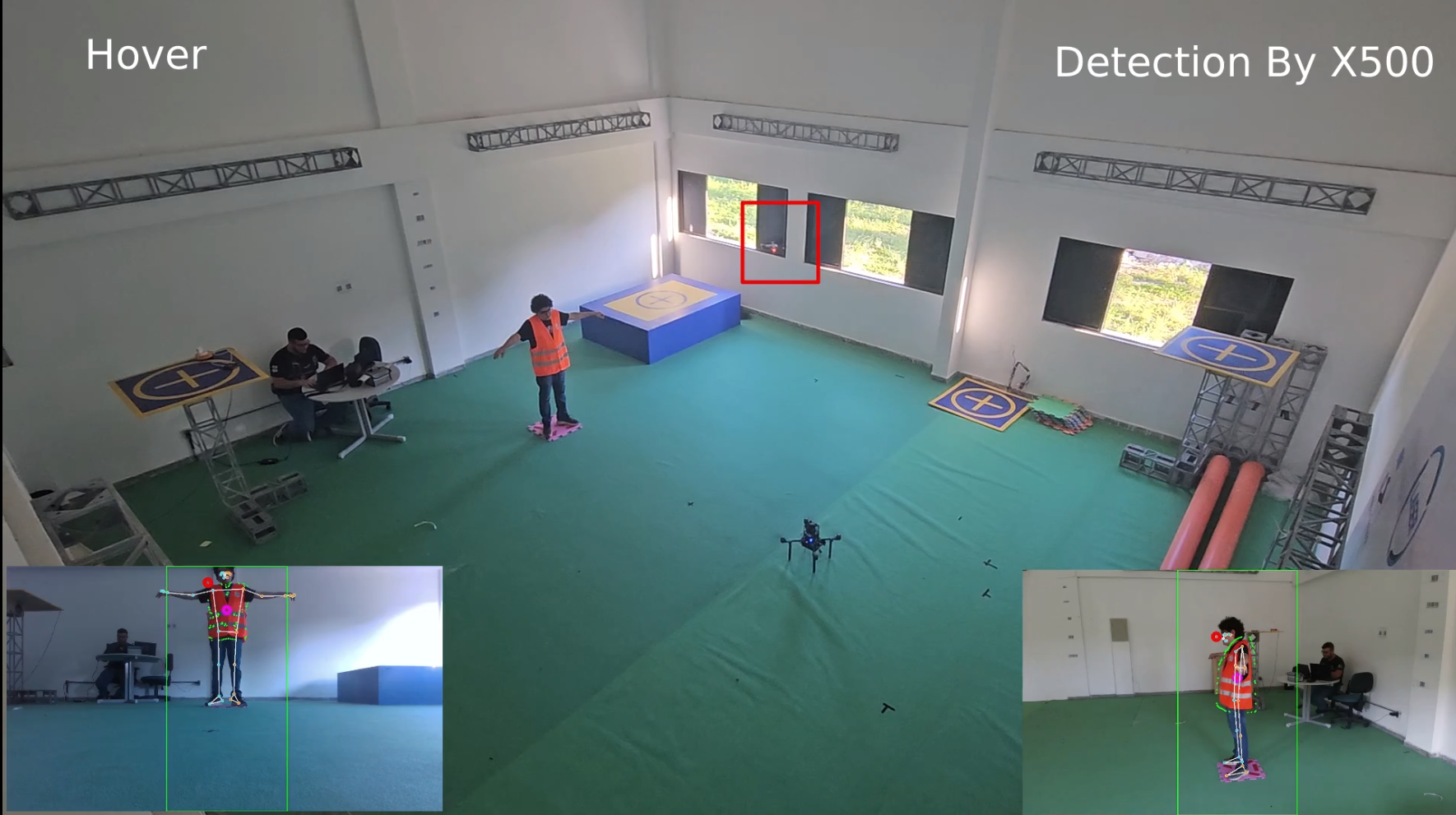}}
    \caption{Drone command action recognition - The lower left corner is the X500 UAV camera, detecting a person making a rover action, the lower right corner is the DJI camera (in red box), detecting a person but not sending any commands because the main control is with the X500. In the top left corner, it is possible to see the action name, and in the top right corner, it is possible to see which UAV detects the person.}
    \label{fig:uavgesture}
\end{figure}

\section{Proposed Pipeline}

In this work, we utilized two different UAVs: a DJI and a x500. The x500 UAV uses a standard control system, the MRS System \cite{baca_mrs_2021} (i.e., a system for UAV flight control and state estimation) as a navigation system. In contrast, the DJI uses only its standard flight controller. The DJI UAV also uses a Real-Time Streaming Protocol (RTSP) server to capture and send the video feed to an offboard computer, which in turn converts it into ROS images. In contrast, the x500 captures the images and converts them into ROS images within its onboard computer. Both the onboard computer of the x500 and the offboard computer use our action recognition algorithm in a distributed fashion (see Fig. \ref{fig:pipe}). In this work, for the sake of simplicity, we assume the DJI is only hovering in order to keep the operator always within the field of view of at least one of the UAVs.

Within our approach, our action recognition has a priority check. This means that if the main drone (i.e., x500 UAV) detects the person, the commands from the secondary pipeline (DJI UAV) are ignored. If not, process the key points to generate the command just like the main pipeline.

\subsection{Gesture Detection}

\begin{figure}[!t]
\centerline{\includegraphics[scale=0.5]{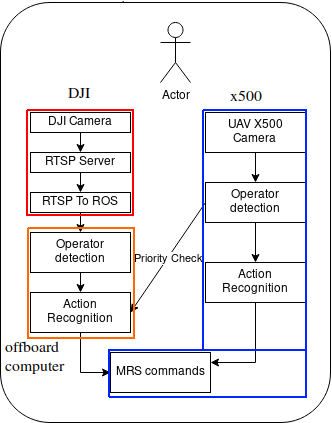}}
\caption{Proposed Pipeline. The DJI UAV (red blocks) sends images to the offboard computer (orange blocks), which runs the action recognition algorithm (LSTM+FL). Meanwhile, an x500 UAV (blue blocks) also captures and processes all the images within the MRS system using our action recognition algorithm (LSTM+FL).}
\label{fig:pipe}
\end{figure}

\begin{figure}[t]
\centerline{\includegraphics[scale=0.5]{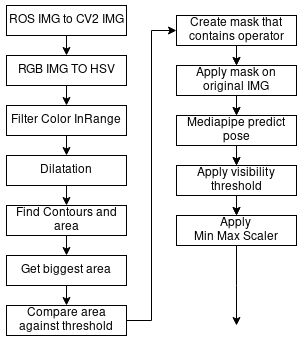}}
\caption{Operator Detection block.}
\label{fig:gesture}
\end{figure}

The \textit{Operator Detection} block within the pipeline (Fig. \ref{fig:pipe}) uses a series of filters (the first nine blocks from Fig. \ref{fig:gesture}) to process the acquired video feed. This is done to isolate the drone operator against persons that can appear on the scene. Then, the filtered images are inserted into the \verb|Mediapipe|\footnote{https://ai.google.dev/edge/mediapipe/solutions/guide}block, which is an open-source framework for building and deploying machine-learning pipelines, and which has a human joint detector. The output of \verb|Mediapipe| is used to calculate the 33 points within the X, Y, and Z coordinates in the world frame. The frame key points are then sent to the action recognition algorithm.

\subsection{Action Recognition}

Our action recognition approach works on a streaming basis reconstructed from each processed frame that comes from the Gesture Detection block. Thus, every time the Action Recognition block receives the position key points from the Operator Detector block, the following operations are performed:

\begin{enumerate}
    \item Creates an array with the first 60 frames received;
    \item Performs the action recognition using the trained LSTM model using FL;
    \item Insert the recognized action into a Moving Average Filter;
    \item Calculate the average of the first three filtered actions;
    \item Sends the action's respective command every 10 seconds.
\end{enumerate}

The actions in our dataset are characterized by the annotated reference points. This information is used to define the architecture of the LSTM we use to classify the actions. The input for the model is defined as $N \times M$, where $N$ denotes the number of frames and $M$ is the multiplication of the number of key points by the number of coordinates. In our experiments, the input for the first LSTM layer has the dimensions of $(60, 99)$ for the used dataset. Therefore, our architecture resulted in a model with six layers, in which the first two are LSTM layers, the third is a dense layer using the ReLU activation function, a third layer is a dropout layer, the fourth layer is another dense layer using the ReLU, and finally a Softmax activation function was added in the last layer. This last layer was responsible for classifying which class the input data belonged to. Finally, the network architecture can be summarized below in Table \ref{table_network_arch}.

\begin{table}[t]
\caption{Architecture Summary}
   \begin{center}
       \begin{tabular}{|c|c|c|c|c|}
       \hline
       \textbf{Layer(type)} & \textbf{Output Shape} & \textbf{Activation Function} \\
       \hline
       LSTM &  30 & ReLU  \\
       \hline
       LSTM &  64 & ReLU  \\
       \hline
       dense\_1 (Dense)  & 64 & ReLU \\
       \hline
       Dropout  & 0.2 & ReLU \\
       \hline
       dense\_2 (Dense)  & 32 & ReLU \\
       \hline
       dense\_3 (Dense)  & 13 & Softmax \\
       \hline
       \end{tabular}
  \label{table_network_arch}
  \end{center}
\end{table}

The third step of our approach is the application of a moving average filter, which is necessary in order to avoid false positive situations that usually happen during transitions between actions, i.e., when the human operator is finishing one action and starting another action. Such situations may happen when the human operator transitions between action $a_1$ and $a_2$ or during the intersection between two complete actions, initiating within the last 30 frames of the first execution and ending at the first 30 frames of the second execution. On the other hand, we also implemented a digital debounce to prevent an overload of commands while the operator sends only one command.

\begin{figure}[!t]
\centerline{\includegraphics[scale=0.25]{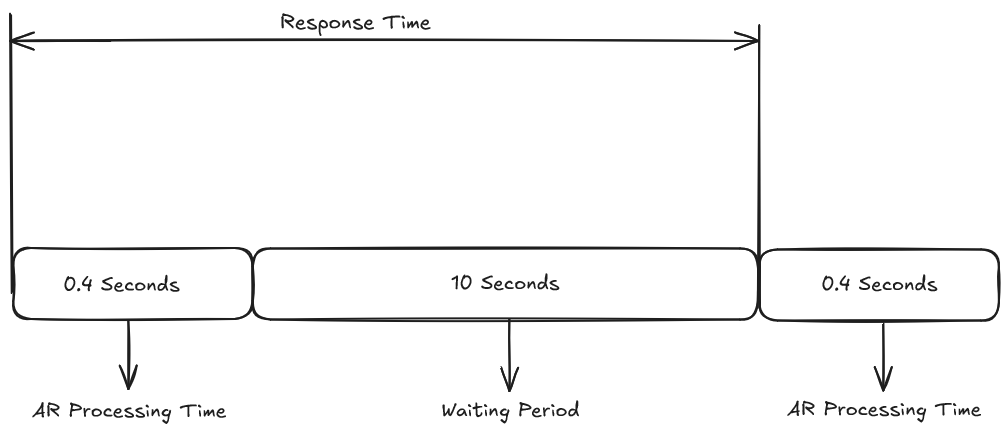}}
\caption{Response Time depiction of 10.4 seconds, where we have a waiting period of 10 seconds and an action recognition (AR) processing time of 0.4 seconds.}
\label{fig:action}
\end{figure}

In resume, our action recognition approach applies a window function considering the average of three actions to avoid detection errors. The entire response time of the Action Recognition approach takes 10.4 seconds, as depicted in Fig. \ref{fig:action}. Then, every recognized action is transformed into a respective command to be sent to the control system. We also apply a ten-second wait window between commands to avoid sending the same command multiple times. Finally, we convert the desired command into control commands that are sent to the x500 UAV MRS system, allowing the UAV to perform the desired actions.

\section{Dataset}

\subsubsection{Dataset from Literature}
Since we focus on the proximal control of drones via action recognition, we opt for a dataset of UAV commanding signals. In this work, we use a dataset created by Perera et al.~\cite{perera_uav-gesture_2019}, which provides \textbf{11 actions} and \textbf{2 static gestures} (i.e., do not contain any limb motion) suitable for basic UAV navigation and command from general aircraft and helicopter handling signals across \textbf{37.151 high-definition video frames} (see Fig. \ref{fig:uavgesture}), all of which are annotated with body joints and action classes. This dataset contains rich variations in the recorded actions in terms of the phase, orientation, camera movement, and body shape of the actors. These properties are a suitable representation of real-world variations and make the dataset extensible to be used in human-robot collaboration environments. Finally, the dataset uses \verb|Openpose|~\cite{cao_openpose_2019} and has 18 reference points to identify a person. Each point has three coordinates, referring to the X, Y, and Z axes. For more details on the dataset, please refer to Perera et al.~\cite{perera_uav-gesture_2019}.

\begin{figure}[ht]
    \centerline{\includegraphics[scale=0.48]{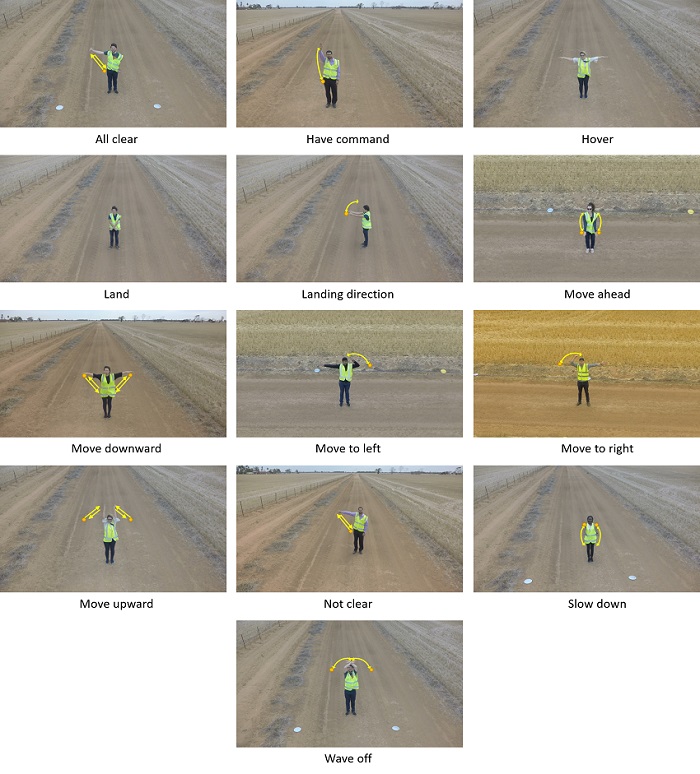}}
    \caption{A frame for each of the eleven actions and two gestures (static) from~\cite{perera_uav-gesture_2019}. The arrows indicate the hand movement directions. The amber color markers roughly designate the start and end positions of the palm for one repetition. The Hover and Land commands are static gestures, i.e., they do not contain active limb motions.}
    \label{fig:uavgesture}
\end{figure}

\subsubsection{Created Dataset}
In addition, we created a second dataset aiming to improve the accuracy of our proposed method\footnote{The Dataset is available at \url{https://github.com/LASER-Robotics/uav_gesture_control_federated_learning}}. Like the UAV Gesture Dataset from Perera et al. \cite{perera_uav-gesture_2019}, our dataset also has \textbf{11 actions} and \textbf{2 static gestures} (i.e., do not contain any limb motion) as commands sent to an operator, a pilot. However, we also have three different points of view of the same action/gesture. Thus, we multiplied the number of gestures/actions, resulting in \textbf{33 unique actions} and \textbf{6 unique static gestures}, with a total of 21.294 video frames. Furthermore, this second dataset also uses \verb|Mediapipe|, with X, Y, and Z coordinates to characterize a subject. A portion of the video (60 frames) and six videos for each action were created to identify the subject. In addition, we also performed a data augmentation operation on key points, a rotation from -15 to 15 degrees in the X, Y, and Z axis. Thus, for the sake of simplicity, let us name it as an Action, both static gestures and moving gestures. Finally, we can formulate mathematically the dataset creation and operations as follows:

\newcommand{\uvec}[1]{\boldsymbol{\hat{{#1}}}}

Data:
\begin{itemize}
    \item  $G$ is the set of actions;
    \item  $V$ is the set of videos for each action;
    \item  $x$ is the frame`s index inside a video.
    \item  $A = \{\uvec{i}, \uvec{j}, \uvec{k}\}$ is the set of vectors.
\end{itemize}

\begin{equation}
    FR_{i,j,x}=R_{v,\alpha} * F_{i,j,x}
    \label{eq:dataset_augmentation}
\end{equation}

Where:
\begin{align}
       & F_{ijx},   \forall  i  \in  G,  \forall  j \in V, \forall  x \in j \label{eq:frame_org}
\end{align}

    \begin{align}
       & R_{v\alpha},  \forall v \in A ;  \alpha \forall \{\alpha \in \mathbb{Z} : -15 \degree \le \alpha \le 15 \degree \} \label{eq:rot_matrix}
    \end{align}
where the $FR$ in  (\ref{eq:dataset_augmentation}) is the resulting frame from the operation between the rotation matrix from a unit vector and an angle in degrees (equation (\ref{eq:rot_matrix})), times the individual frame in each video from each action (equation (\ref{eq:frame_org})).

\section{Federated Learning-based Action Recognition in UAVs}

In this section, to help mitigate the aforementioned issues related to the control of UAVs in human-robot collaborative scenarios, especially with respect to the training process, we detail our FL-based (FL-based) proposal. Let us begin considering several UAVs ($u_i$) in a multi-robot system ($\mathcal{U}$), where $u_i \in \mathcal{U}$, $i \in \{1\ldots N\}$ and $N$ is the total number of robots. Let us also consider the existence of a global dataset ($\mathcal{D}$) that describes multiple instances of the actions $a$ that we target, where $a \in \mathcal{D}$. Thus, in this work, we let each UAV $u_i$ own a possibly overlapping partition of the global dataset $\mathcal{D}$. In other words, for each UAV $u_{i} \in \mathcal{U}$ we associate a partition $d_{i}$ such that $\cup_{i=1}^{\infty} d_{i} = \mathcal{D}$. Naturally, we admit the possibility that $d_{i} \cap d_{j} \neq 0$ for any arbitrary pair of UAVs, $u_{i}$ and $u_{j}$. Following the FL protocol, we use a central server to organize the model inference on $\mathcal{D}$, using the $N$ available datasets. The process starts with the server, which sends a copy of the initial model parameter values, $\boldsymbol{w}_{init}$, for each robot $u_i$. Then, each robot updates its local model, $\phi_i$,  using its own dataset $d_{i}$. Following that, each updated local model parameter $\boldsymbol{w}_{i}$ is sent back to the server where they are aggregated -- often via an averaging operation -- into a new $\boldsymbol{w}_{init}$. The process continues until convergence criteria are reached. Note that in this setup, no actual $d_{i}$ is transmitted to the central server, but only the model parameters $\boldsymbol{w}_{i}$.

Recall that the motivation for considering an isolated partition $d_{i}$ is related to the constraint imposed by the operation of each $u_{i}$. These devices can operate in isolation, avoiding the exposure to personal data via each $d_{i}$, while sharing their unique experiences in the field via $\boldsymbol{w}_{i}$. Additionally, each device can update its model on the fly, which is beneficial, especially in case the available computing power does not meet the demand imposed by the time complexities involved.

\begin{figure}[!t]
\centerline{\includegraphics[scale=0.4]{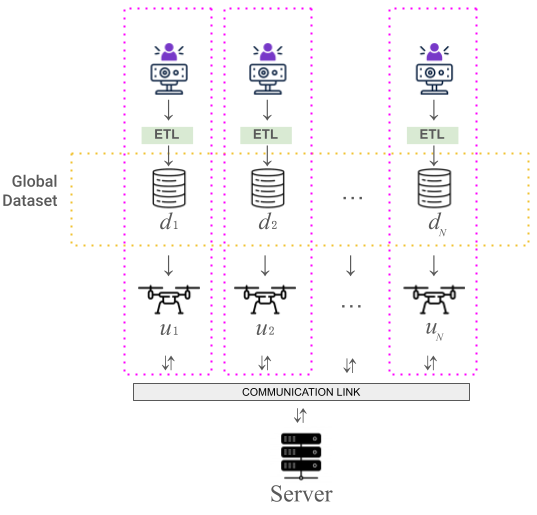}}
\caption{Training process using FL. A central server aggregates model parameters from $N$ models trained on each robot's $u_{i}$ local data $d_i$. The model architecture is assumed to be constant across all robots. Via individual ETL processes, each $u_{i}$ will contribute to learning the global dataset $\mathcal{D}$ without sharing the user's data. The global model inference happens only through the model parameters exchanged via the aggregation operation on the server side.}
\label{fig:my_arch}
\end{figure}

\begin{figure}[!t]
\centerline{\includegraphics[scale=0.5]{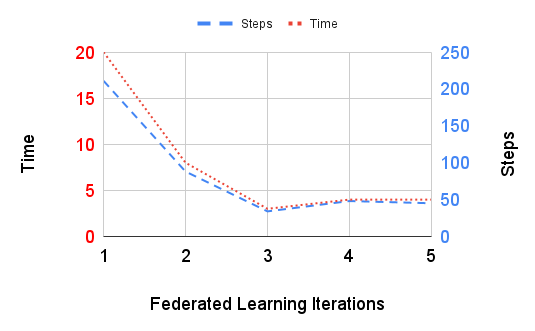}}
\caption{Training results of two UAVs.}
\label{fig:training}
\end{figure}

\begin{algorithm}[!t]
\begin{algorithmic}
\caption{Federated Average and the LSTM implementation. The $K$ clients are indexed by $k$; $B$ is the local minibatch size, $E$ is the number of local epochs, and $\eta$ is the learning rate}
\label{alg:main_project}

    \Procedure{Server Executes}{$a,b$}
        \State initialize $w_0$
        \For{each round t = 1, 2, ...}
            \State $m\gets max(C*K, 1)$
            \State $S_t \gets$ (random set of $m$ clients)
            \For{each client $k \in S_t$ \textbf{in parallel}}
                \State $w_{t+1}^k\gets $ ClientTrainingLSTM($k$, $w_t$)
            \EndFor
            \State $m_t   \gets \sum_{k \in S_t} n_k$
            \State $w_{t+1} \gets \sum_{k \in S_t} \frac{n_k}{m_t} w_{t+1}^k$
        \EndFor
        \EndProcedure
        
        \Procedure{ClientTrainingLSTM}{$k,w$} 
        \State $\mathcal{B} \gets$ (StratifiedShuffleSplit $P_k$ into batches of size $B$)
        \For{each local epoch $i$ from $i$ to $E$}
            \For{batch $b \in \mathcal{B}$}
                \State $w\gets w - \eta \nabla \ell(w;b)$
                \State $\eta \gets $ CallbackReduceLROnPlateau($\eta$, loss)
                \State CallbackEarlyStoppingAndModelSave($w$)
            \EndFor
        \EndFor
        \State return $w$ to server
        \EndProcedure

        \Procedure{CallbackEarlyStoppingAndModelSave}{$w$} 
        \State // Save the best weight and stop the training
        \EndProcedure
        
        \Procedure{CallbackReduceLROnPlateau}{$\eta$,loss} 
        \State // Reduce learning rate when a metric has stopped improving.
        \State return $\eta$ to client
        \EndProcedure

\end{algorithmic}
\end{algorithm}

To capture the patterns involved in the categorization of each action $a$, we pose the training of an LSTM model via the FL framework. Such a process returns a converged LSTM model over the dataset $\mathcal{D}$ represented by each UAV partition $d_{i}$. The training cycles start with each UAV $u_{i}$ training their model on their data $d_{i}$ after receiving the initial parameters from the server. This training process happens through scheduled rounds of parameter learning, as shown in Fig. \ref{fig:my_arch}. Naturally, this implies that communication between the devices is established. Although outside the scope of our work, this can also be extended to defined cross-swarm training cycles to allow maximum adaptation of individual swarms to patterns that are cross-regional w.r.t their operation. Algorithm \ref{alg:main_project} details the standard federate average process for LSTM within the \textit{Server block} of Fig. \ref{fig:my_arch}. In this work, we used only two UAVs. Thus, we can see in Fig. \ref{fig:training} that our system only needs three interactions of the FL process for the training since no significant improvement would have been achieved if more federated iterations had been executed.

\section{Results}

Our experiments and simulations relied on \verb|Flower|\cite{beutel_flower_2022}, a wildly popular framework for FL and from which we used the federated average parameter aggregation strategy described in Algorithm~\ref{alg:main_project}. The computation was performed by a portable computer with an i7 processor, equipped with an Nvidia graphics card, running Ubuntu LTS and ROS Noetic. 

\begin{figure}[t!]
\centering
\includegraphics[width=0.9\linewidth, frame]{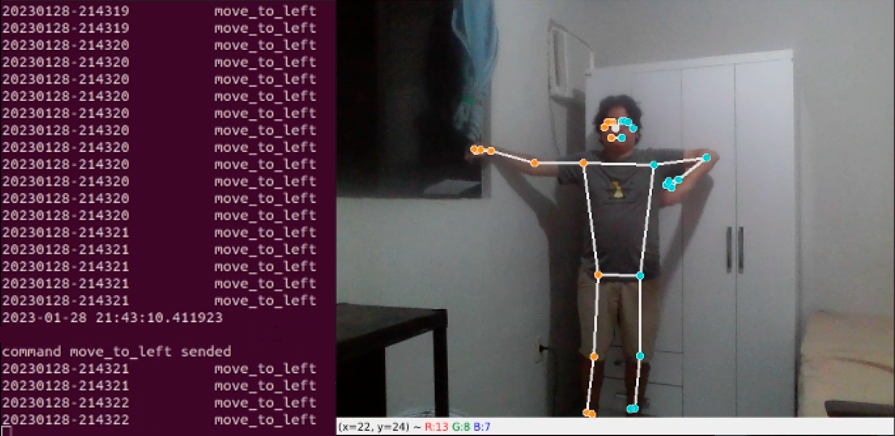}
\caption{Real-time acquisition. Client A detects and sends the correct command to the UAV. }
\label{fig:client_a_command}
\end{figure} 

\begin{figure}[t!]
\centering
\includegraphics[width=0.9\linewidth]{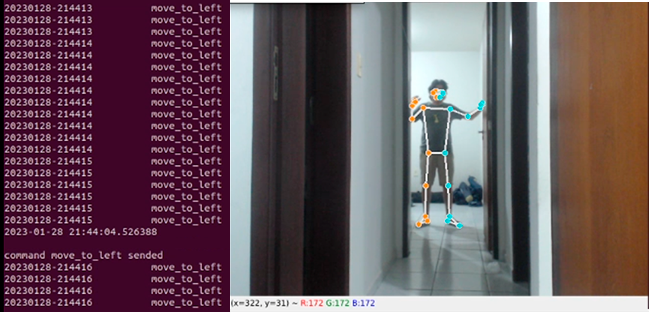}
\caption{Real-time acquisition:  Client B detects and sends the wrong command to the UAV. }
\label{fig:client_b_command}
\end{figure} 

\begin{figure}[t!]
\centering
\includegraphics[width=0.9\linewidth]{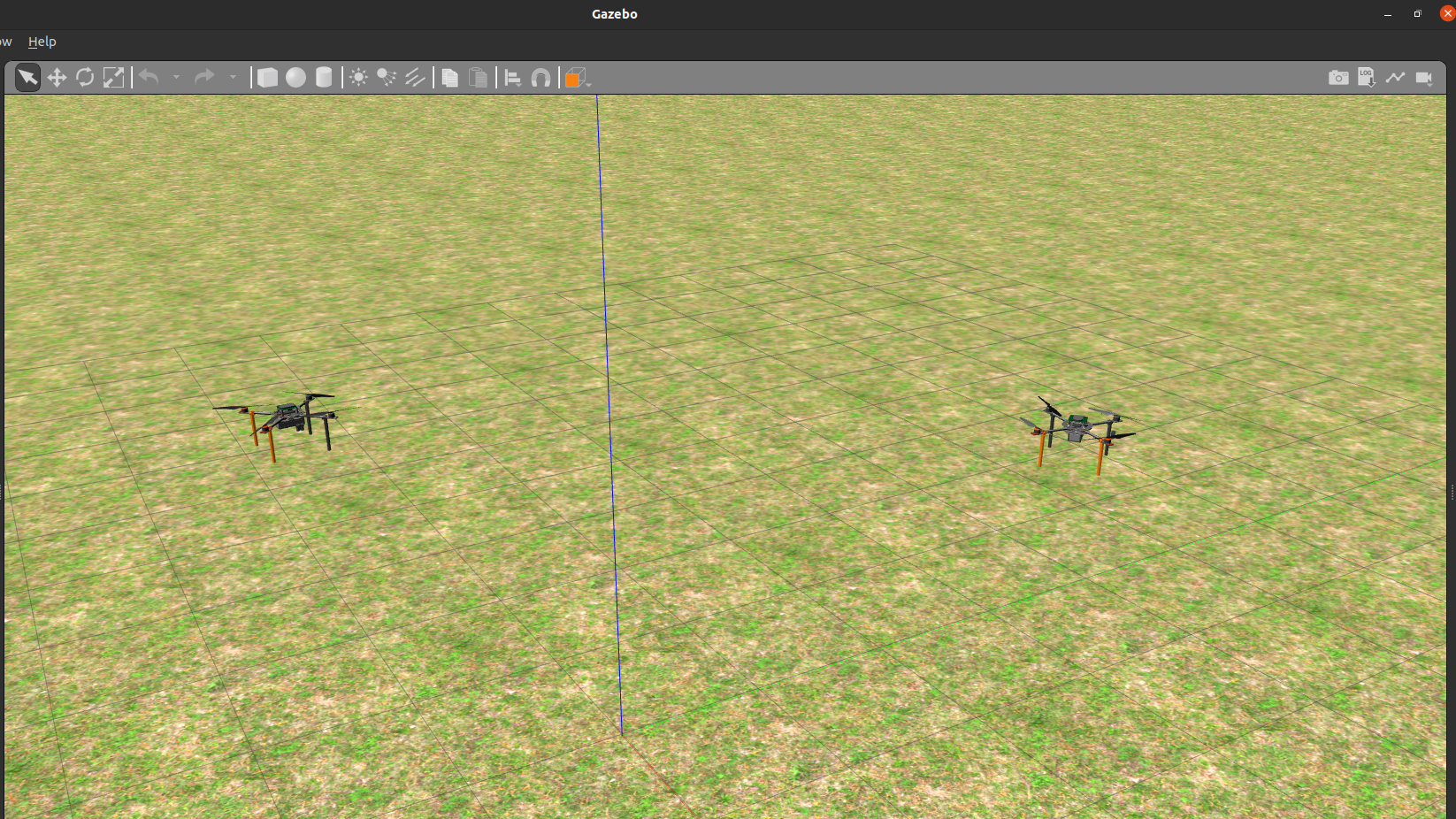}
\caption{Simulated UAVs.}
\label{fig:gazebo}
\end{figure}

\subsection{Realistic Simulations}

\begin{table}[t!]
\caption{Summary of metrics - Dataset 1 - Client A and B}
    \begin{center}
        \begin{tabular}{|c|c|c|c|c|}
        \hline
        \textbf{Client} & \textbf{Loss}     & \textbf{Accuracy} & \textbf{Time elapsed (seconds)} \\
        \hline
        A               & 2.29              & 0.99               &  235  \\
        \hline
        B	            & 3.08              & 0.75	            &  233  \\
        \hline
        \end{tabular}
    \label{tb:dataset_1_a_b}
    \end{center}
\end{table}

\begin{figure}[t!]
    \centering
    \includegraphics[width=0.9\linewidth]{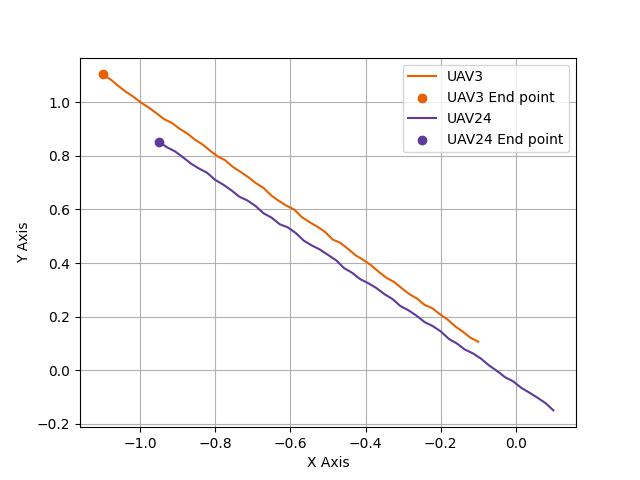}
    \caption{The flight path from both UAVs. }
    \label{fig:Flight_Path}
\end{figure} 

\begin{table}[t]
\caption{Summary of metrics - Dataset 2 - Client A, B, C}
    \begin{center}
        \begin{tabular}{|c|c|c|c|c|}
        \hline
        \textbf{Iteration}     & \textbf{Loss}   & \textbf{Accuracy} & \textbf{Time Elapsed (s)}\\ 
        \hline
        A                      & 0.12            & 0.96     &    934 \\
        \hline
        B                      & 0.04            & 0.97     &    934 \\
        \hline
        C                      & 0.07            & 0.97     &    934 \\ 
        \hline
        \end{tabular}
    \label{tb:dataset_3_a_b_c}
    \end{center}
\end{table}

The performed simulations aimed to evaluate the accuracy and time elapsed or our proposed architecture and model when using each of the abovementioned datasets. The best training parameters were found empirically by using MinMaxScalar from Scikit-Learn \cite{scikit-learn} with 750 epochs and an Early Stopping of 300 with five iterations of federated average on the server. 

\begin{figure*}[!t]
\centering
\subfigure[Hover command.\label{fig:hover}]{\includegraphics[width=2.5in]{images/Hover.png}}
\hfil
\subfigure[Have command.\label{fig:have_command}]{\includegraphics[width=2.5in]{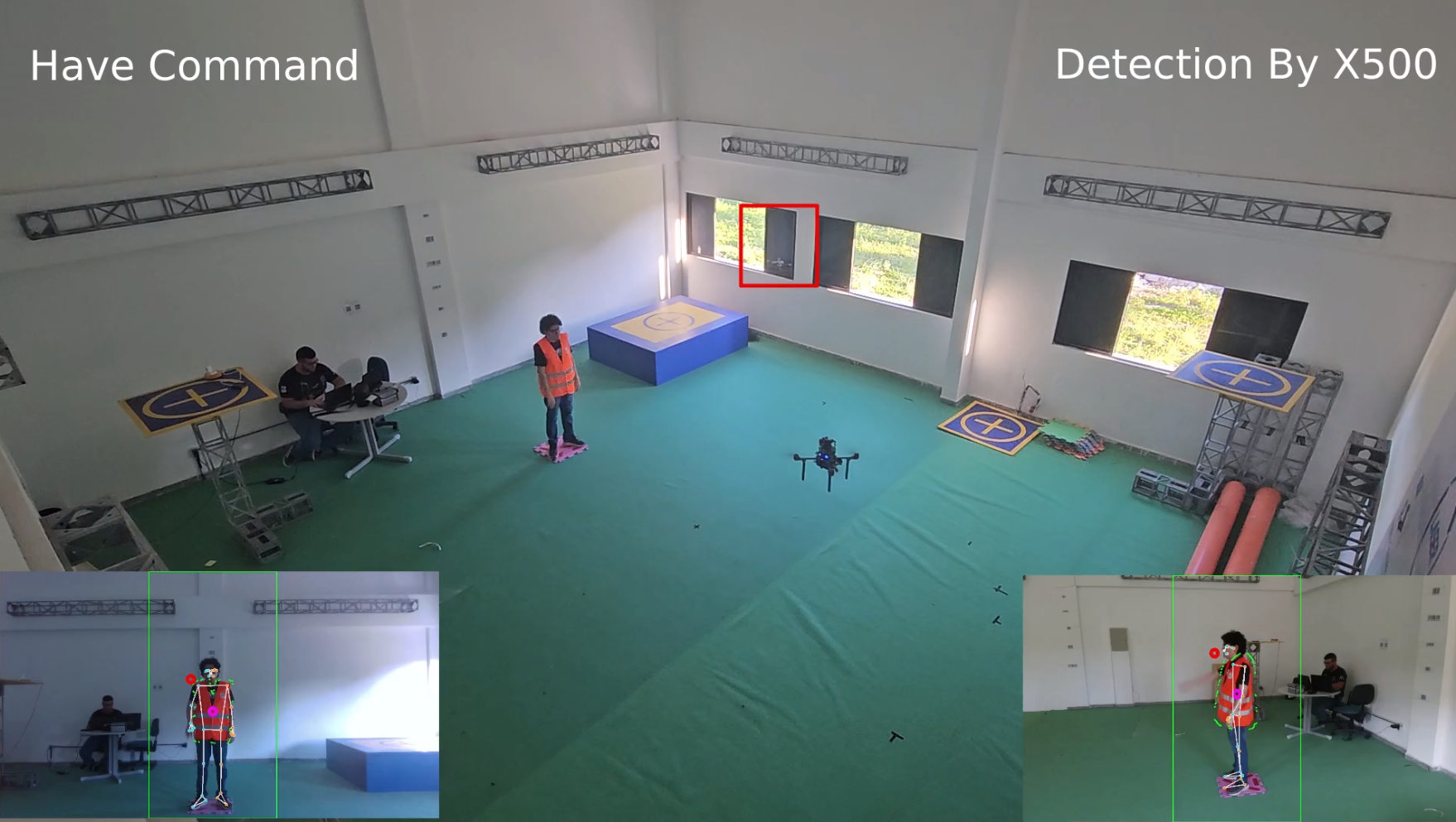}}
\hfil
\subfigure[Move to left command. \label{fig:move_left}]{\includegraphics[width=2.5in]{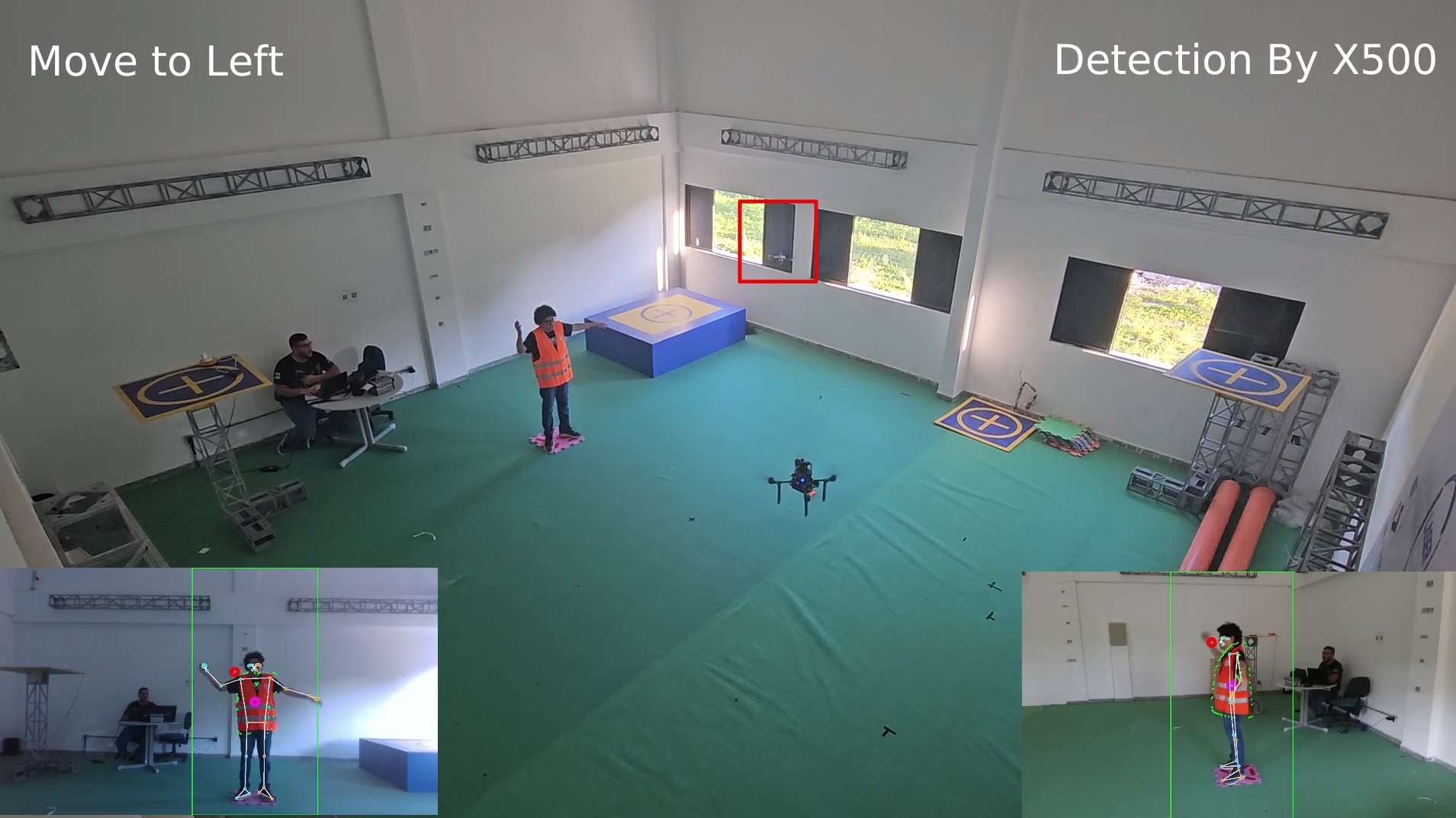}}
\hfil
\subfigure[Have command action has been detected. \label{fig:enter-label}]{\includegraphics[width=2.7in]{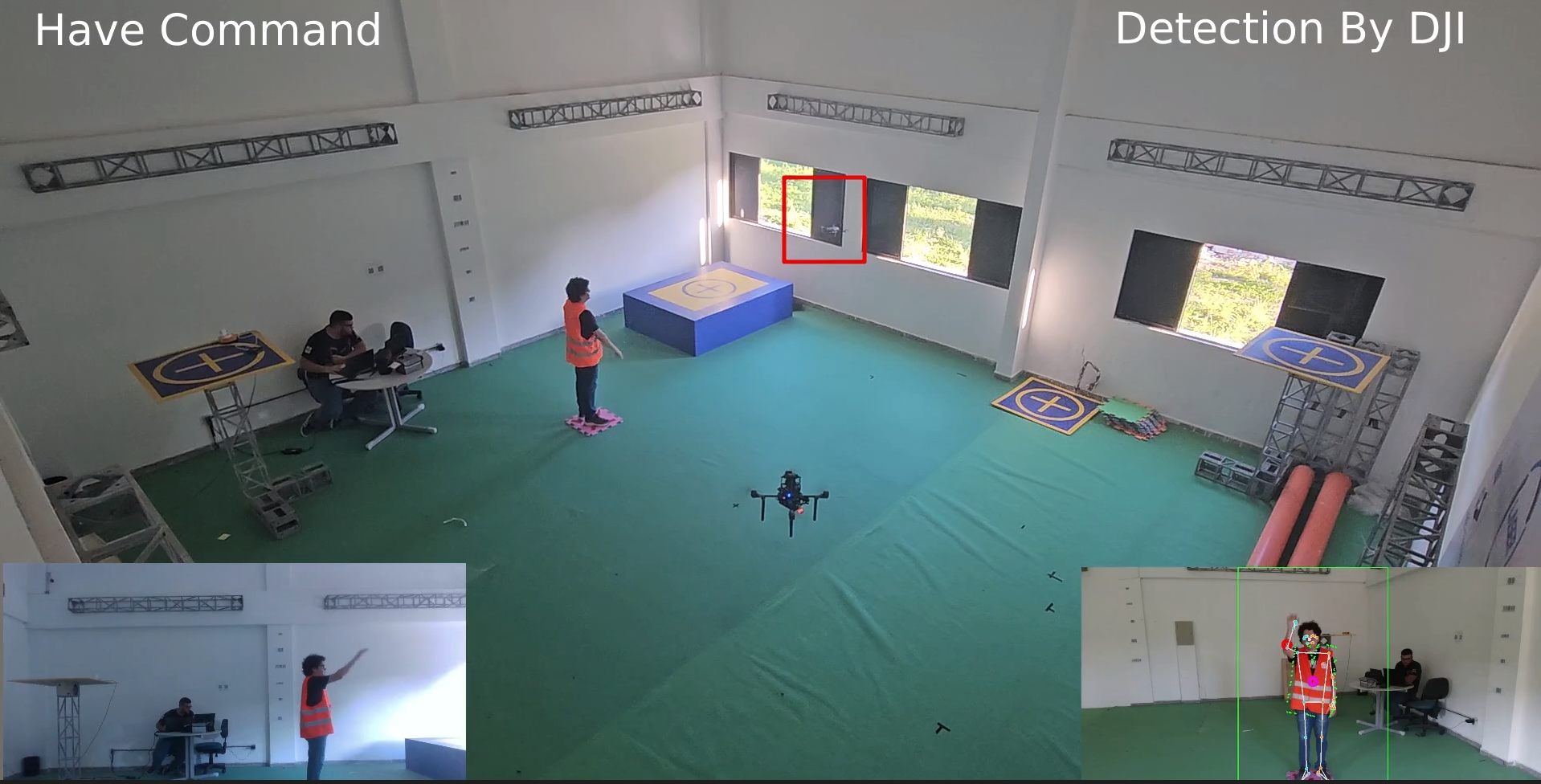}}
\caption{Human-UAV interaction experiment.}
\label{fig:exp_comparison}
\end{figure*}

\begin{figure}[t]
    \centering
    \includegraphics[width=0.8\linewidth, frame]{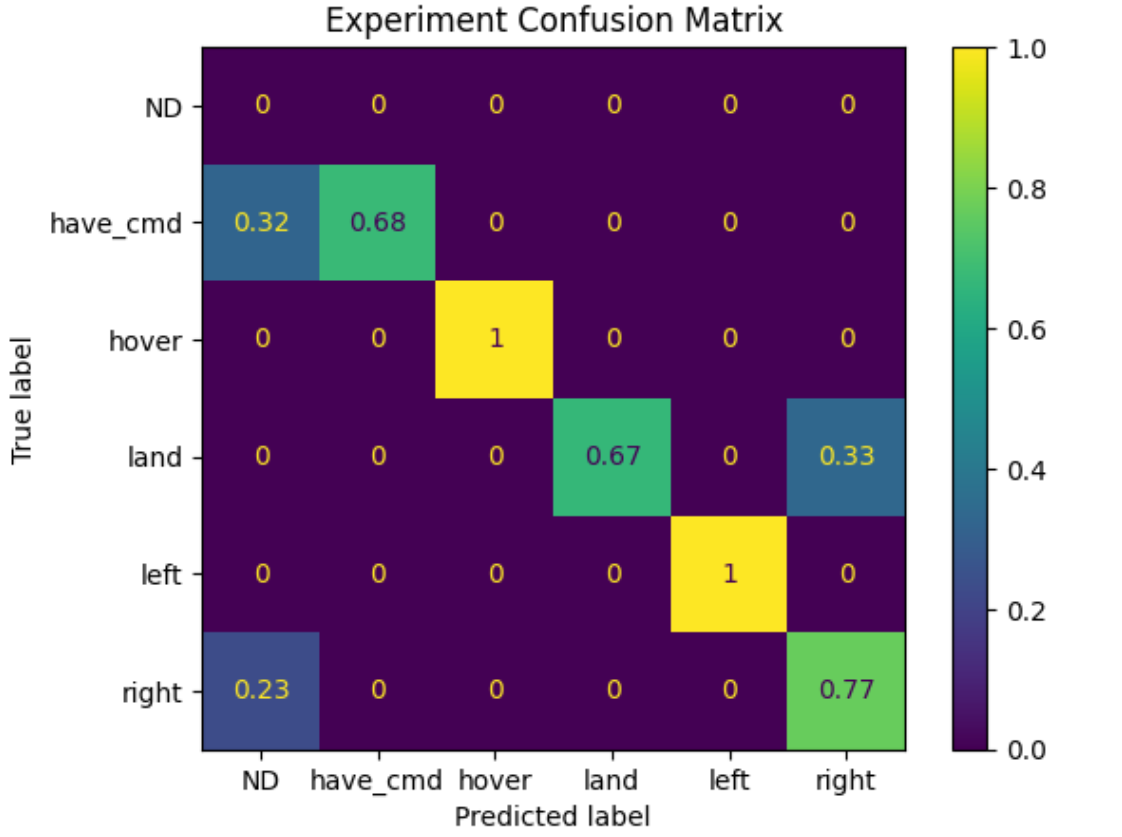}
    \caption{Resulting confusion matrix from the experiments.}
    \label{fig:confusion}
\end{figure}

The simulations used dataset 1 \cite{perera_uav-gesture_2019}  to validate the model after the training. With dataset 1, we analyzed the case where we have only two visual observation points (i.e., client A and client B), where each client had the proposed architecture implemented, using the same dataset for both clients.

In these simulations, each client was associated with a different camera and also a different field of view. The actions from the dataset were semantically connected to UAV commands. Fig. \ref{fig:client_a_command} and Fig. \ref{fig:client_b_command} present examples of the simulations carried out, with the acquisition of images from two cameras in two different positions detecting the operator's actions and sending commands to two simulated UAVs. Fig. \ref{fig:gazebo} shows the realistic simulation of the UAVs.

The performance of the neural network was measured in terms of accuracy. Although the accuracy during the training procedure was satisfactory, the validation results were different from the expected. The evaluation values for loss and accuracy can be seen in Table \ref{tb:dataset_1_a_b}. With an accuracy of 99\% (client A), the simulated UAV was able to detect the person and send the correct command, as is shown in Fig. \ref{fig:client_a_command}.

Although the accuracy of the second UAV was 75\% (client B), the simulated UAV was also able to detect the person and send the correct command, as is shown in Fig. \ref{fig:client_b_command}. Then, these detected commands triggered the UAV-1 drone corresponding to client A towards its final position (with coordinates (-1.1;1.1)) and the UAV-2 drone corresponding to client B towards its final position (with coordinates (-0.95;0.90)). The final path towards the desired destination point can be seen in Fig. \ref{fig:Flight_Path}. 

\subsection{Real-robot Experiments}

For real robot experiments, we used three clients to test the trained model, representing three visual observation points: client A, client B, and client C. In real robot experiments, it is expected to have a drop in accuracy in human action detection. Therefore, we needed to improve the dataset used by creating a second dataset and re-training the architecture. The simulations performed to validate the trained model improved significantly. The \textbf{results from the model validation} are in Table \ref{tb:dataset_3_a_b_c}. The time elapsed compared to the other datasets increased, but also did the detection accuracy, with all the clients learning with an end-around accuracy of 97\%.

With these validation results, we performed real robot experiments with dataset 2 and two clients, client A, an X500 UAV, and client B, a DJI UAV. The main UAV (i.e., the X500 UAV) used the MRS System \cite{baca_mrs_2021} to fly, whereas the DJI used its benchmark software. A video of this experiment can be seen at \url{https://youtu.be/AO01r8JV5fw}. In these experiments (see Fig. \ref{fig:exp_comparison}), the objective was to command the X500 UAV to perform movements based on the actions performed by the human operator. Consequently, by performing desired actions, the X500 UAV moves to a position where the human operator cannot be visualized. Meanwhile, the DJI UAV is flying in a fixed position, always having the human operator in its field of view. This configuration allows our FL-based architecture to always detect the actions performed by the operator.

The human operator performed the following commands in order: (X500) UAV Hover (Fig. \ref{fig:hover}), Have Command (Fig. \ref{fig:have_command}), Move to Left ((Fig. \ref{fig:move_left})), (DJI) Have Command (Fig. \ref{fig:enter-label}), Move to Right, Land (Low detection by X500), Land (Detected by DJI, landed).
For the commands to be translated into position commands, the operator must perform a main command action (i.e., Have Command) to turn on the recognition system and send control commands to the drone. After detecting the Have Command action, the operator can now perform the Move to Left command.

Once the X500 UAV moves, the human is visually partially blocked (e.g., the feet of the human cannot be seen anymore), and thus, the detection fails. To overcome this problem, the human operator turns to the DJI UAV, enabling the second drone to visualize the human fully. Fig. \ref{fig:enter-label} shows that it was possible to see that the DJI realized the detection and later sent it to the UAV X500. With the human within the field of view of the DJI, the commands Move to Right and Land were successfully sent to the X500 UAV. Finally, it is important to mention that this code can also support more drones or cameras to control one or more drones. 

We can also state the absence of overfitting since there is a difference between a wrong action prediction and no action detected. In the first case, the person is detected by the UAV, but the predicted action is wrong. To avoid this problem, we used the moving average filter. The second case is when the UAV is unable to detect the operator at all, leading to our system also being unable to predict the action. \textbf{In the real-time experiments, the Accuracy of our system was 0.6867}, with the confusion matrix shown in Fig.\ref{fig:confusion}, formed by comparing the true action (label) vs. the predicted action. This drop in accuracy is explained by several sources of uncertainties within the robotic system (e.g., distance from the operator, quality of the sensor, changes in light conditions, etc.). Finally, the measured \textbf{precision of our system was 0.93} while the calculated \textbf{F1 score was 0.83}.

\section{Conclusions}

This work proposes a Long Short-Term Memory (LSTM) Deep Neural Networks-based action control detector composed of two layers in association with three densely connected layers and FL for multi-UAV proximal control. This approach relies on visual detection performed by the UAVs with onboard cameras. We demonstrated that the proposed architecture performs with a precision of 93\% (similar to the state-of-the-art for a benchmark dataset UAV-Gesture \cite{perera_uav-gesture_2019} and for our own proposed dataset), reaching acceptable accuracy values during training and evaluation, making it possible to control a drone with actions safely. This improvement was achieved by pre-processing the video before the prediction of the machine learning algorithm and removing destructive noises that cause outliers. Finally, the model was tuned with more use cases and possible noises that improve the algorithm performance and validated with real robot experiments. In future works, we aim to improve the robustness of our approach, validating it with experiments within complex environments with different lighting conditions and wind.

\bibliographystyle{IEEEtran}
\bibliography{root}

\end{document}